\pdfoutput=1

\documentclass[11pt]{article}

\usepackage[]{acl}
\usepackage{booktabs}    
\usepackage{times}
\usepackage{latexsym}
\usepackage{colortbl}
\usepackage{longtable}
\usepackage{supertabular}
\usepackage{amsmath}
\usepackage{xspace}
\usepackage[T1]{fontenc}
\usepackage[utf8]{inputenc}
\usepackage{arydshln}
\usepackage{wasysym}
\usepackage{microtype}

\usepackage{inconsolata}

\usepackage{graphicx}

\usepackage{multirow}

\usepackage{xcolor}

\newcommand{\revise}[1]{\textcolor{black}{#1}}

\newcommand{\redchapter}[1]{\begingroup\color{black}#1\endgroup}

\title{\ours: Targeted Synthetic Data Generation for Practical Reasoning over Structured Data}

\author{Xiang Huang\textsuperscript{1}, Jiayu Shen\textsuperscript{1}\thanks{~~Equal contribution.}, Shanshan Huang\textsuperscript{1}, Sitao Cheng\textsuperscript{2}, \\ \textbf{Xiaxia Wang\textsuperscript{3}, Yuzhong Qu\textsuperscript{1}} \\ 
\textsuperscript{1}State Key Laboratory for Novel Software Technology, Nanjing University, China \\
\textsuperscript{2}University of California, Santa Barbara \\
\textsuperscript{3}University of Oxford \\
xianghuang@smail.nju.edu.cn, yzqu@nju.edu.cn}

\begin{document}
\newcommand{\ours}{\textsc{Targa}\xspace} 

\maketitle

\begin{abstract}
Semantic parsing, which converts natural language questions into logic forms, 
plays a crucial role in reasoning within structured environments. 
However, existing methods encounter two significant challenges: reliance on extensive manually annotated datasets and limited generalization capability to unseen examples. 
To tackle these issues, we propose Targeted Synthetic Data Generation (\ours), a practical framework that dynamically generates high-relevance synthetic data without manual annotation. 
Starting from the pertinent entities and relations of a given question, we probe for the potential relevant queries through layer-wise expansion and cross-layer combination. 
Then we generate corresponding natural language questions for these constructed queries to jointly serve as the synthetic demonstrations for in-context learning.
Experiments on multiple knowledge base question answering (KBQA)  datasets demonstrate that \ours, using only a 7B-parameter model, substantially outperforms existing non-fine-tuned methods that utilize close-sourced model, achieving notable improvements in F1 scores on GrailQA~(+7.7) and KBQA-Agent~(+12.2).
Furthermore, \ours~also exhibits superior sample efficiency, robustness, and generalization capabilities under non-I.I.D. settings.

\end{abstract}

\section{Introduction}

Reasoning over structured environments, such as Knowledge Base~(KB), Database, and Web, has emerged as a crucial ability for large language models (LLMs)~\cite{liu2023agentbench,gu2023dont}.
Among various methods for structural reasoning, semantic parsing stands out as a mainstream and has garnered increasing attention from researchers.  
By translating natural language questions~(NLQ) into logic forms, semantic parsing enables seamless interaction with structured environments, thereby enhancing user experience and accessibility.

\begin{figure}
    \setlength{\belowcaptionskip}{-0.5cm}
    \centering
    \includegraphics[scale=1]{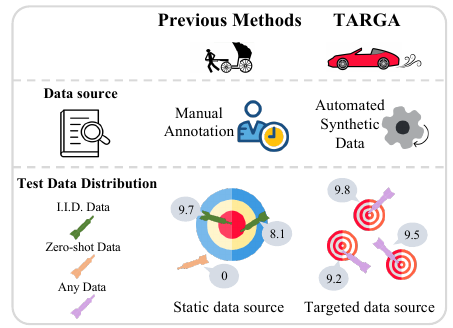}
    \caption{Compared with previous methods, 
    \ours aims to mitigate the reliance on large amounts of manually labeled data and enhance generalization capabilities in non-i.i.d. scenarios.
    }
    \label{fig:comparison} 
\end{figure}

However, current semantic paring methods typically face two significant challenges:  

\textit{1) Dependence on annotation}. 
% Previous methods
Previous methods usually rely on extensive amounts of manually annotated data.
By training~\cite{Ye2021rng,shu2022tiara} or retrieval~\cite{li2023few,nie2023codestyle} based on large-scale annotations, existing works have made remarkable progress.
Unfortunately, collecting manual annotations in specific environments is labor-intensive and time-consuming.
In real-world scenarios, large amounts of pre-collected annotated data are often unavailable, limiting the scalability of these methods.

\textit{2) Limited generalization capability}. 
Even with access to large annotated datasets, previous methods still struggle with generalizing to unseen examples.
Regardless of the paradigm (\textit{e.g.,} in-context learning or fine-tuning), any method relying on a static, offline-collected dataset is inevitably influenced by the dataset's distribution.
Specifically, these methods tend to perform well on examples encountered in the dataset (the I.I.D. setting) but exhibit weaker generalization when faced with unseen environmental items or query structures (the non-I.I.D. settings), as shown in Figure \ref{fig:comparison}. 
In complex environments, such as Freebase~\cite{bollacker2008freebase} with over three billion triples, it is nearly impossible for a pre-collected static dataset to cover the full scope of the environment. Additionally, as the coverage of annotations increases, so do the costs associated with training and retrieval, which further limits the scalability and generalization. 

To address the aforementioned challenges, in this work, 
we propose a practical semantic parsing framework called \underline{Tar}geted Synthetic Data \underline{G}ener\underline{a}tion (\textbf{\ours}), which does not need any manually annotated data and can efficiently work on a 7B model.
% To tackle the above challenges, 
Specifically, \ours~addresses these challenges by dynamically synthesizing highly relevant examples of a test question as demonstrations for in-context learning.
Starting from the KB items~(entity, relation) that may be related to the given question,
we construct logic forms through layer-wise expansion~(extend a new edge for a sub-structure) and cross-layer combination~(combine different sub-structures), gradually evolving from simple to complex structures.
To further enhance relevance, we re-rank the synthetic logic forms to select the most pertinent ones and generate their corresponding natural language questions, which are then used as demonstrations for reasoning. 
Through this automatic data synthesis, \ours~free annotators from the heavy burden of labeling tasks.
Additionally, the demonstrations are generated based on the given question, thus naturally avoiding the challenge of generalization.

Without any data annotation, \ours significantly outperforms all non-fine-tuned approaches across multiple complex KBQA datasets, particularly excelling in non-I.I.D. settings. 
Remarkably, \ours achieves this with only a 7B-parameter model, whereas most baselines rely on advanced closed-source models, such as \textit{gpt-3.5-turbo}, enabling faster and more cost-efficient inference.
On the GrailQA dataset, we improve the performance of non-fine-tuned methods from an F1 score of 61.3 to 69.0.
On KBQA-Agent, the most challenging dataset, we elevate the SOTA performance from 34.3 to 46.5 F1 scores.
Further analyses highlight the high quality of the data generated by \ours. 
Even with a single demonstration, \ours still surpasses all non-fine-tuned methods on GrailQA.
Additionally, \ours exhibits remarkable robustness in adversarial settings\footnote{\url{https://github.com/cdhx/TARGA}}.

\section{Related Works}
\subsection{Few-shot KBQA with LLMs}

With the advancement of large language models, recent works have adopted LLMs as the backend for KBQA.
In particular, In-Context Learning~\cite{brown2020language} requires dozens of demonstrations to guide the model's responses. 
To achieve competitive performance, existing ICL-based KBQA works~\cite{li2023few,nie2023codestyle} typically retrieve the most similar examples from a manually annotated training set as demonstrations.
However, this strategy often results in performance degradation on non-I.I.D. questions involving unseen structures or KB items. 
For example, \citet{nie2023codestyle,li2023few} reported that the performance in zero-shot settings can be up to 20\% lower compared to I.I.D. settings.

Another line of KBQA methods, agent-based methods~\cite{liu2023agentbench,huang2024queryagent,gu2024middleware}, decomposes questions into individual steps 
% for an agent 
to solve.
While step-by-step solving aligns with human intuition and demonstrates remarkable generalization ability, it incurs high computational costs and presents challenges in constructing trajectories.
Moreover, the effectiveness of the agent-based paradigm relies heavily on the planning and generalization abilities of advanced LLMs, leading to subpar performance when using weaker models, such as some open-source variants.
Such dependency underscores the limitation of agent-based approaches when superior LLMs are unavailable or impractical to use due to resource constraints.

\subsection{Synthetic Data Generation}
Instead of relying solely on human annotation for training data, recent works have leveraged LLMs to generate synthetic data, thereby reducing the burden on human annotators. 
For instance, \citet{vicuna2023,alpaca} use instructions to generate training data as a supplement to manual annotation via self-instruct techniques \cite{wang-etal-2023-self-instruct}.
However, this approach still requires human-annotated seed examples to ensure high-quality demonstrations, which entails significant demand for LLM usage.
Rather than directly prompting LLMs to generate training data, \citet{cao2022kqa,huang2023markqa} address this problem by first sampling structured queries from the environment and then converting these queries into natural language using LLMs.
Nevertheless, obtaining meaningful structured queries remains a non-trivial task.

Other works similar to ours include FlexKBQA~\cite{li2023flexkbqa} and BYOKG~\cite{agarwal2024bring}.
FlexKBQA relies on predefined templates and model training stages to automatically annotate data, while BYOKG synthesizes data from scratch. 
% but still needed a manual template  and 
However, both approaches require a time-consuming offline data collection phase.
More importantly, like other methods, they rely on reasoning over a pre-collected static dataset, which still suffers from generalization issues.
In \ours, we systematically design a framework to dynamically synthesize relevant examples in an online manner.
It avoids the need for a lengthy data collection process, enabling us to dynamically obtain the most relevant examples for each test case without being constrained by a static dataset.

\begin{figure*}
    \setlength{\abovecaptionskip}{0.2cm}
    \centering
    \includegraphics[scale=0.84]{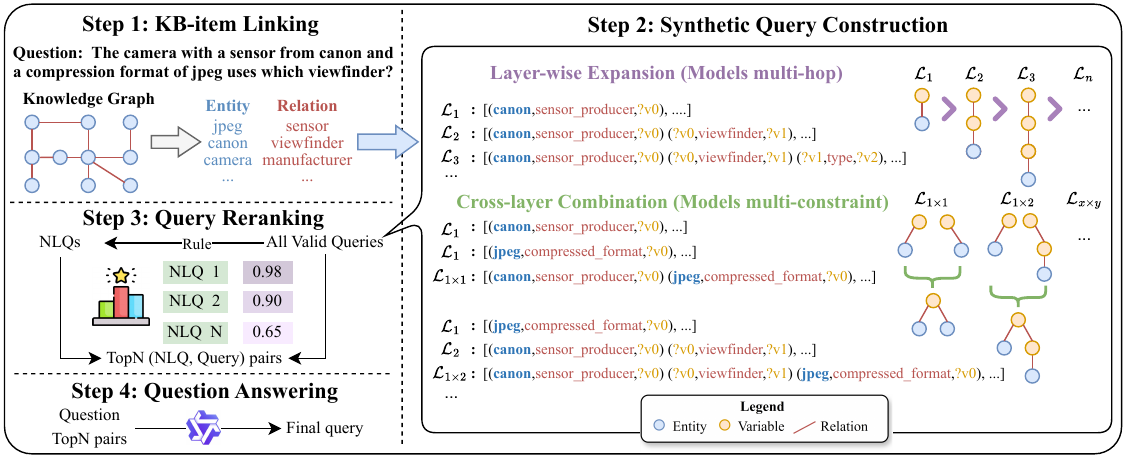}
    \caption{Overview of \ours.  
    % Given a natural language question, we first collect candidate KB items such as entity and relation as initialization. 
    % Then, we explore valid query graphs, expanding these items from simple to complex structures to construct synthetic queries. 
    % Next, we filter high-relevance candidate queries by ranking and generating their corresponding natural language questions. 
    % Finally, we use these high-quality synthetic data as demonstrations for QA reasoning.
    % Our \ours~framework enables the automatic generation of targeted data by synthesizing queries relevant to the question. We achieve this by gradually expanding retrieved items from simple to more complex structures. Without manual annotation, we still ensure generalizability by maintaining query relevance via re-ranking.
    }
    \label{fig:framework}
\end{figure*}

\section{Methods}

\subsection{Overview}

As shown in Figure \ref{fig:framework}, our targeted data synthesis framework, \ours, consists of four parts.
\revise{Given a natural language question $nlq$, we first collect candidate KB items such as entities $E_{nlq}$ and relations $R_{nlq}$ as initialization. 
Then, we explore valid query graphs from simple to complex structures to construct synthetic queries $Q$. }
Next, we filter high-relevance candidate queries by ranking. Finally, we use these high-quality synthetic data as demonstrations for QA reasoning.

\subsection{Candidate KB Items Retrieval}
\label{subsec:kb_items_retrieval}
 
For candidate entities, we adopt the linking result provided by~\citet{gu2023dont} for a fair comparison~(detail in Appendix \ref{subsec:entity_linking_detail}).
For candidate relations, we compute the similarity of question and freebase relations based on \textit{text-embedding-ada-002}\footnote{\url{https://platform.openai.com/docs/guides/embeddings}} and retain the top 20 most similar candidates.
Different from previous fine-tuned methods~\cite{Ye2021rng,hu2022logical}, which typically require training a relation linking model for higher precision and recall, 
our method does not rely on the precision of these items.
This is because, although all retrieved KB items are relevant to the question, they do not necessarily form valid combinations within a specific graph structure. 
In this way, the subsequent query construction steps in Section~\ref{subsec:syhthetic_query_generation} can be viewed as a joint entity-relation disambiguation process, thus significantly reducing the number of invalid queries.

\redchapter{
\subsection{Synthetic Query Construction}
\label{subsec:syhthetic_query_generation}

This stage aims to construct question-targeted queries to facilitate subsequent QA reasoning. 
Given a knowledge base $\mathcal{G}$ and the set of retrieved KB items relevant to the question $nlq$,
we explore the possible query structures $Q$ that are valid~(\textit{i.e.,} yield non-empty execution results). 
However, enumerating all possible structures may lead to an unmanageable combinatorial explosion.
To mitigate this, our exploration of candidate queries follows a simple-to-complex manner, where we only further explore new structures that are derived from the sub-structures already verified as valid.
% that have been verified as valid.
Starting from the simplest structure~($\mathcal{L}_1$ in Figure \ref{fig:framework}), we progressively search for more complex query structures through \textit{Layer-wise Expansion}~(for multi-hop structures) and \textit{Cross-layer Combination}~(for multi-constraint structures), gradually extending the obtained query graphs until a desired complexity is achieved.

\textbf{Layer-wise Expansion}
is utilized to model multi-hop structures (the depth of the query graph), which are chain-like, non-branching query structures originating from a single entity. 
We define $\mathcal{L}_{k}$ as the set of queries in which the distance from the entity to the farthest variable in the chain-like query structure is $k$.
Specifically, we first identify all possible connections between $E_{nlq}$ and $R_{nlq}$, forming the simplest query structure $\mathcal{L}_{1}$, where an entity $s$ is connected to a variable $o$ through a single relation $p$. 
\textsc{Exec}\big($q$, $\mathcal{G}$\big) indicates the execution results of query $q$ against $\mathcal{G}$.

\vspace{-10pt} 
\begin{equation}
\begin{split}
\mathcal{L}_{1} = \big\{(s, p, o) \mid & s \in E_{nlq}, p \in R_{nlq}, \\
& \textsc{Exec}\big((s, p, o), \mathcal{G}\big) \neq \emptyset \big\}.
\end{split}
\end{equation}
\vspace{-8pt}

We then progressively expand outward from the terminal variable nodes by connecting them to a new variable through another relation to construct $\mathcal{L}_{2}$, and so forth.
Generally, $\mathcal{L}_{k+1}$ is formed by expanding the valid queries from the previous layer ($\mathcal{L}_{k}$) with an additional edge:

\vspace{-10pt} 
\begin{equation}
\begin{split}
\mathcal{L}_{k+1} = \big\{ q \cup (o_i, p', o_j) \;\big|\; 
 q \in \mathcal{L}_{k}, \;  o_i \in \mathcal{O}(q),
& \\ p' \in R_{nlq}, \textsc{Exec}\big(q \cup (o_i, p', o_j), \mathcal{G}\big) \neq \emptyset \big\}.
\end{split}
\end{equation}
\vspace{-10pt}

where $\mathcal{O}(q)$ represents the set of variables in $q$ and $o_j$ is a newly introduced variable.
The expansion process stops when the complexity threshold (e.g., 3 hops) is reached, since for a coherent and reasonable question, the distance between a specific entity and the final answer typically does not exceed three hops.
 
\textbf{Cross-layer Combination}
models multi-constraint structures (the width of the query graph) by merging two queries, thereby applying multiple constraints to the same variable. 
Given two queries $q$ and $q'$, we choose one of the variables from each query~($o_i$ for $q$ and $o_j$ for $q'$) as the common variable of them, then combine these two queries into a more complex query through this shared variable.
We define $\mathcal{L}_{x \times y}$ as the set of queries formed by combining a query from  $\mathcal{L}_{x}$ and a query from $\mathcal{L}_{y}$.  
Specifically, we start from the simplest combinations, such as merging two queries in $\mathcal{L}_{1}$, and gradually explore more complex combination patterns, such as merging $\mathcal{L}_{2}$ with $\mathcal{L}_{3}$
or merging $\mathcal{L}_{1}$ with $\mathcal{L}_{1 \times 2}$.
This combination process can be formally expressed as:
\begin{equation}
\begin{aligned}
\mathcal{L}_{x \times y} = \big\{ q \cup q' \;\big|\;
q \in \mathcal{L}_{x}, \; q' \in \mathcal{L}_{y}, & \\
\exists o_i \in \mathcal{O}(q), \; o_j \in \mathcal{O}(q'), 
 & \\ \mathcal{E}(o_i) \cap \mathcal{E}(o_j) \neq \emptyset, 
 \textsc{Exec}\big(q \cup q', \mathcal{G}\big) \neq \emptyset \big\}\,, &
\end{aligned}
\end{equation}
where $\mathcal{E}(o_i)$ refers to the set of entities corresponding to the variable $o_i$ in the execution result of query $q$ on $\mathcal{G}$.
% where $\mathcal{E}(o)$ denotes the set of entities that the variable $o$ represent in $\mathcal{G}$, 
$o_i$ and $o_j$ serve as the shared variable.
This combination terminates once the query structure reaches five edges, which is sufficient to model most questions in current datasets. 
}

In this manner, we circumvented a significant number of invalid queries, thus obtaining most of the potentially relevant queries with relatively lower query overhead. 
\revise{We also provide statistical data regarding these synthetic queries in Appendix \ref{appendix:statics_of_query_construction}.}
% As shown in Table~\ref{tab:query_construct}, 
The average number of valid candidate queries per question is only in the range of several dozen, which is well within the contextual length limits manageable by an ICL model. 

\subsection{Synthetic Query Re-ranking}
\label{subsec:reranking}
To obtain the most relevant examples for the subsequent QA task, we re-rank all valid queries using the \textit{bge-reranker-v2-m3} model~\cite{chen2024bge} based on their similarity to the question.
Additionally, we employ a process called Query Textification, where the synthesized query is transformed into a format closer to natural language through heuristic rules.
This step helps bridge the gap between the text embedding model and the query, further improving the quality of the ranking.
\revise{Detail and examples of textification process are provided in Appendix~\ref{appendix:textification} and Table \ref{tab:textification}.}

To address the imbalance caused by the exponential growth of complex queries, we implement a Hierarchical Ranking strategy. 
\redchapter{For all queries derived from the same parent query~(the sub-query that this query is derived from), we retain only the top $n$ candidates. The final candidate query pool is the union of all top tanked candidates:
\begin{equation}
Q_{\text{ranked}} = \bigcup_{a \in \mathcal{A}} 
\underset{q \in Q, \, \textsc{Parent}(q) = a}{\textsc{Argmax}^{(n)}} \textsc{Score}(QT(q), nlq).
\end{equation}
where $Q$ denotes the set of queries generated during query construction, 
$\textsc{Score}$ measures similarity and $QT$ is Query Textification.
% $\te t'hextsc{Score}(TQ(q), nlq)$ represents the similarity between a query $q$ and the question $nlq$, 
$\textsc{Parent}(q)$ indicates the parent query of $q$, and $\mathcal{A}$ refers to the set of parent queries that have child queries.
This approach ensures that the size of the candidate pool grows at a manageable rate, while preserving high-quality queries for downstream processing.
}

\begin{table*}[!t]
    \setlength{\belowcaptionskip}{-0.4cm}
    \centering
    \resizebox{0.97\textwidth}{!}{
    \begin{tabular}{lccccc}
    \toprule
    \textbf{Methods} & \textbf{Models} & \textbf{GrailQA} & \textbf{GraphQ} & \textbf{KBQA-Agent} & \textbf{MetaQA} \\
    \midrule
    \rowcolor{gray!20} \multicolumn{6}{c}{\textit{\textbf{full} training set (Seq2Seq Fine-tuning / ICL / Agent Fine-tuning)}} \\
    ArcaneQA~\cite{gu2022arcaneqa} & T5-base & 73.7 & 31.8 & - & - \\
    Pangu~\cite{gu2023dont} & T5-3B & 83.4 & 57.7 & - & - \\
    \hdashline
    KB-Binder-R~\cite{li2023few} & GPT-3.5-turbo & 58.5 & 32.5 & - & 99.5 \\
    KB-Coder-R~\cite{nie2023codestyle} & GPT-3.5-turbo & 61.3 & 36.6 & - & - \\
    \hdashline
    KG-Agent~\cite{jiang2024kgagent} & Llama2-7B & 86.1 & - & - & - \\
    DARA*~\cite{fang2024dara} & Llama2-7B & 77.7 & 62.7 & - & - \\
    \midrule
    \rowcolor{gray!20}\multicolumn{6}{c}{\textit{\textbf{dozens} of annotations~(ICL)}} \\
    KB-Binder~\cite{li2023few} & GPT-3.5-turbo & 50.8 & 34.5 & 4.2 & 96.4 \\
    KB-Coder~\cite{nie2023codestyle} & GPT-3.5-turbo & 51.7 & 35.8 & - & - \\
    Pangu (ICL)~\cite{li2023few} & Codex & 53.5 & 35.4 & 18.1 & - \\
    \midrule
    \rowcolor{gray!20}\multicolumn{6}{c}{\textit{\textbf{one} annotation~(Agent Training-free)}} \\
    AgentBench~\cite{liu2023agentbench} & GPT-3.5-turbo & 30.5 & 25.1 & 25.9 & - \\
    FUXI~\cite{gu2024middleware} & GPT-3.5-turbo & - & - & 34.3 & - \\
    QueryAgent~\cite{huang2024queryagent} & GPT-3.5-turbo & 60.5 & 50.8 & - & 98.5 \\
    \midrule
    \rowcolor{gray!20} \multicolumn{6}{c}{\textit{\textbf{zero} annotation~(ICL)}} \\
    BYOKG~\cite{agarwal2024bring} & MPT-7B & 46.5 & - & - & 56.5 \\
    \textbf{\ours (Ours)} & QWen-2.5-7B-Instruct & 69.0 & 50.6 & 46.5 & 85.7 \\
    & QWen-2.5-72B-Instruct & \textbf{70.6} & \textbf{54.1} & \textbf{57.3} & 99.8 \\
    & GPT-3.5-turbo & 68.9 & 51.0 & 52.7 & 96.5 \\
    & GPT-4-turbo & 69.8 & 52.5 & 51.4 & \textbf{99.9} \\
    \bottomrule
    \end{tabular}
    }
    \caption{Main results of KBQA performance, categorized by the amount of required annotated data example. Seq2Seq Fine-tuning / ICL / Agent Fine-tuning indicates different reasoning paradigms~(split by the dash line). \textbf{Bold} values highlight the best among non-fine-tuned models. * indicates using golden entity linking result.}
    \label{tab:main_result}
\end{table*}

\subsection{Question Answering}
To help the LLM understand the semantics of the provided query, we equip each generated query with its corresponding natural language questions~(NLQ), forming (NLQ, Query) pairs. 
Specifically, we directly utilize the textification results mentioned in Section \ref{subsec:reranking} as the NLQ, ensuring both efficiency and the preservation of information integrity. 
Then, we adopt the In-context Learning paradigm to generate the target query. 
Finally, we parse and execute the output query from the LLM to obtain the answer.

\section{Evaluation}
 
\subsection{Setup}
We experiment with four complex KBQA datasets, \textit{i.e.,} GrailQA \cite{gu2021beyond}, GraphQ \cite{su2016graphquestions}, KBQA-Agent \cite{gu2024middleware}, MetaQA \cite{zhang2017variational} and a Text2SQL dataset, \textit{i.e.,} WikiSQL \cite{zhong2017Seq2SQL}. 
We use the F1 scores as the evaluation metric for KBQA and denotation accuracy for Text2SQL.
We compare \ours with various paradigms of baselines, including fine-tuning, ICL, and Agent, where we report performance in the original paper.
For experiments with other settings, we copy the re-implemented result from \citet{gu2024middleware}.
By default, we use \texttt{Qwen-2.5-7B-Instruct} as the base LLM in our experiments with 10 demonstrations for all datasets.
Detailed introduction of datasets and baselines are available in Appendices~\ref{appendix:datasets} and \ref{appendix:baselines}.
 
\subsection{Main Result}
Table \ref{tab:main_result} illustrates the main result for KBQA. we compare \ours with methods that require different amounts of annotation.
% As shown in Table \ref{tab:main_result}, 
For the relatively challenging datasets, \textit{i.e.,} GrailQA, GraphQ, and KBQA-Agent, based on a 7B model, \ours~achieves the best performance among all non-fine-tuned methods which are based on advanced close-sourced LLMs.  
% and even surpass some fine-tuning methods on GrailQA and GraphQ.
On GrailQA and KBQA-Agent, \ours~surpasses previous SOTA non-fine-tuned methods by 8.7 and 13.0 F1.
On GraphQ, \ours~even beats some fine-tuned methods and achieves similar performance with the best non-fine-tuned method.
% On KBQA-Agent, our method elevate the best performance from 34.3 to 52.7.

When compared to methods with a similar paradigm~(ICL-based),
\ours~outperform previous methods by 14.0 and 28.4 in F1 on GraphQ and KBQA-Agent, respectively.
% the advantage on GraphQ and KBQA-Agent widens to a substantial margin of 14.0 and 28.4 points.
It is worth noting that our method requires neither any manually annotated corpus nor the expensive close-sourced model.
Besides, we have not incorporated self-consistency to boost the performance.
This can be attributed to the high quality of the synthetic data, which has led to a reduction in task difficulty and a decreased reliance on the capabilities of strong LLMs.
Moreover, compared with other ICL-based methods which include 40-100 demonstrations, \ours~uses only 10 demonstrations but still achieves the best performance, demonstrating notable data efficiency.

Compared to BYOKG which also works without annotated data, \ours achieves approximately 1.5$\times$ performance on GrailQA and MetaQA-3Hop.
More importantly, \ours~dynamically synthesizes the most relevant data for different questions, enabling seamless adaptation to questions from any distribution.
Besides, the synthetic data by \ours~is generated online, eliminating the need for a time-consuming offline data collection phase.

\begin{figure}[t!]
    \setlength{\belowcaptionskip}{-0.4cm}
    \centering
    \includegraphics[width=0.9\linewidth]{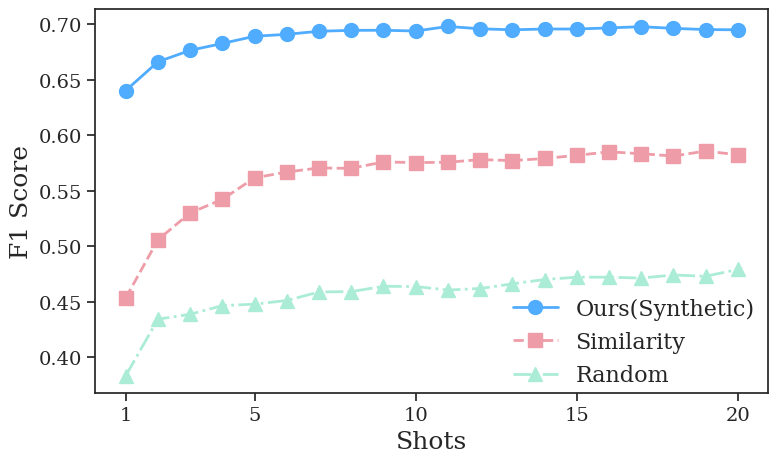}
    \caption{Performance with various numbers of demonstrations on GrailQA (1,000 randomly sampled questions). 
    % The random setting indicates the use of randomly extracted examples from the training set as the demonstrations while the similarity setting uses the most similar example.
    }
    \label{fig:different_shots}
\end{figure}

\subsection{Detailed Analyses}
% ICL是否有question，question和lf是否匹配，对结果的影响
To gain more insights, we conduct detailed experiments to illustrate some favorable practical characteristics of~\ours on: sample efficiency, robustness, generalization ability, efficiency, model size requirements, and transferability.

\subsubsection{Sample Efficiency}
\label{subsec：hyper_param}
In this section, we analyze how the number of demonstrations impacts the performance.
We experiment on GrailQA, with the number of demonstrations ranging from 1 to 20. 
Based on our QA framework, we compare three distinct sampling settings: 
\textbf{Random}, \textbf{Similarity}-based, and \textbf{Ours~(synthetic)}, corresponding to examples randomly sampled from the training set, retrieved by similarity from the training set, and retrieved by similarity from the synthetic data by \ours, respectively.
The random and similarity settings can be viewed as reflections of the previous ICL-based and the retrieval-augmented ICL-based methods.
Results are illustrated in Figure \ref{fig:different_shots}. 
% Our findings can be summarized in two points. 
With only one demonstration, our synthetic setting significantly outperforms the random and retrieval settings with 20 shots, suggesting the high quality of our synthetic data.
% This indicates the high quality of our synthetic data.
Moreover, the growth curve in the synthetic setting (blue line) is relatively flat as the number of demonstrations increases.
% also suggesting the quality of our synthetic method.
After reaching 7 shots, the synthetic setting exhibits almost no further improvement, while the other two settings continue to show growth even after reaching 20 shots, highlighting the data efficiency of our methods.

\subsubsection{Robustness Analyses}
\label{subsec:robustness_analyze}

\begin{figure}[t!]
    \setlength{\belowcaptionskip}{-0.4cm}
    \centering
    \includegraphics[width=0.9\linewidth]{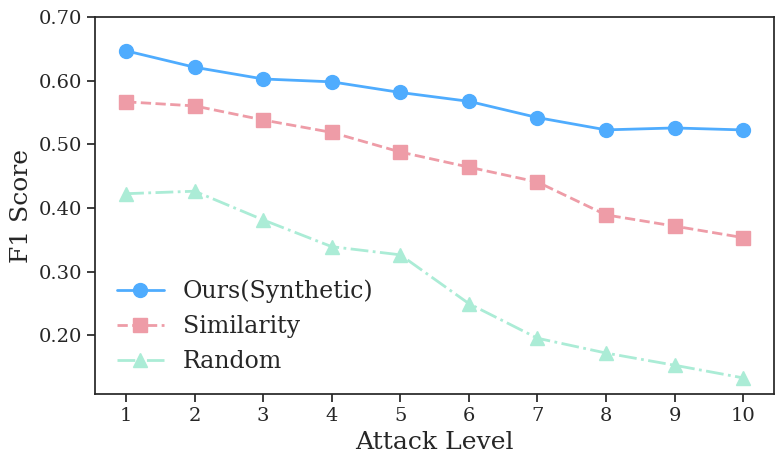}
    \caption{Performance under attack setting on 1,000 randomly sampled GrailQA questions. Attack level indicates how many demonstrations have been corrupted.}
    \label{fig:attack}
\end{figure}

To further validate the robustness of our approach in real-world scenarios, we conduct an adversarial experiment designed to simulate conditions of poor synthetic data quality. 
Specifically, the attack involves randomly replacing one relation in a candidate query. 
We compare the same three settings as in Section \ref{subsec：hyper_param}.
As in Figure \ref{fig:attack}, our method exhibits significantly stronger robustness under adversarial conditions.
Even when all demonstrations were compromised, the performance degradation of \ours~was only around 25\%.
% the performance degradation did not exceed 13 points, 
In contrast, the other setups experience a sharp decline: the F1 scores of similarity-based setup drop by about 40\%, and the random setting even falls by approximately 75\%.
This further demonstrates the superior robustness of our method compared to other approaches. 
We also provide another analysis about when corrupt entities in demonstrations in Appendix \ref{appendix:robust_entity}

\subsubsection{Performance on Different Generalization Levels}
% Fuse设定的实验，iid数据在训练集中一模一样和只有实体不一样的分布
% @sjy 表格 

We experiment on GrailQA to compare the performance on different generalization levels~(preliminary in Appendix~\ref{appendix:generalization_level}).
To make a fair comparison with the ``-R'' setting of previous methods, we also implemented a ``-R'' version of \ours where the entire training set was incorporated into our demonstration sampling pool. 
Specifically, we retain the top 5 most similar examples from the training set and the top 5 most similar synthetic data instances jointly as demonstrations. 

As shown in Table~\ref{tab:generalization}, under the I.I.D setting, \ours-R benefits from the inclusion of high-quality annotations~(training set),
% achieving noticeable improvements and 
achieving similar performance to previous ``-R'' methods.
However, for the more challenging Compositional and Zero-shot settings, where similar questions are absent from the pre-collected training set, the performance of previous methods in the ``-R'' setup dramatically decreases by approximately 30 in F1. 
In contrast, \ours shows no significant decline, demonstrating its strong generalization ability in scenarios that more closely resemble real-world situations where relevant corpora are unavailable.
Notably, under the zero-shot setting, \ours derives minimal improvements from the training set, suggesting that pre-collecting a substantial corpus of examples is ultimately an incomplete solution and tends to fail when confronted with real, unseen problems.

\begin{table}[t]
\centering  
\resizebox{0.95\columnwidth}{!}{
    \begin{tabular}{lcccc}
    \toprule
    \textbf{Methods} & \textbf{I.I.D.}  & \textbf{Comp.} & \textbf{Zero.}  & \textbf{Avg.} \\
    \midrule          
    \textsc{KB-Binder} &  48.3  & 48.8  & 41.8 & 50.8 \\
    \textsc{KB-Coder} & 49.3   & 49.6   & 43.2 & 51.7\\
    {\ours} & \textbf{68.4}   &  \textbf{62.2} & \textbf{71.7}  & \textbf{69.0} \\
    \midrule
    \textsc{KB-Binder-R} &  80.6  & 53.6  & 50.7  & 58.5 \\
    \textsc{KB-Coder-R} & \textbf{81.0}   & 57.8   &  54.1 & 61.3\\
    % \textsc{~~ w/ Training set } & 80.9  & 65.3 &  70.1 & 71.6\\
    \textsc{\ours-R } & 80.8 & \textbf{63.6} &  \textbf{71.6} & \textbf{71.9}\\
     \bottomrule   
    \end{tabular} 
}
\caption{Results of different generalization levels on GrailQA.
``-R'' indicates a version accessing the whole training set for similarity retrieval.
Comp. and Zero. indicates the Compositional and Zero-shot setting on GrailQA, respectively. Avg. denotes the average F1.}
\label{tab:generalization}
\end{table}

\subsubsection{Efficiency analysis}
\label{subsec:efficiency_analysis}
For a practical QA system, high efficiency is also a key characteristic.
Following \citet{huang2024queryagent}, we analyzed three efficiency metrics: TPQ, QPQ, and CPQ, as shown in Table~\ref{tab:efficiency}.
Regarding TPQ, our method significantly outperforms previous methods with only 4.5 seconds response time on GrailQA~(\revise{detailed in Appendix~\ref{appendix:runtime_breakdown}}). 
Regarding QPQ, Agent-based methods have an inherent advantage. 
However, comparatively speaking, the overhead of QPQ remains relatively inexpensive than the other two metrics. 
Compared to KB-Binder, which also employs the ICL paradigm, our approach demonstrates a marked superiority on QPQ.
This is primarily because our synthetic demonstrations are highly aligned with the target question, enabling the generated logic forms to be executable without any post-processing in most cases.  
Conversely, since previous methods can not always retrieve the relevant candidate query in the training set, the generated logic forms are often not executable.
Consequently, the logic form necessitates stepwise binding to valid KB items, which leads to a large demand for queries.

In terms of CPQ, agent-based methods inherently face challenges due to the lengthy trajectory of demonstrations and the need for multiple calls of LLM.
Since our method does not rely on close-sourced LLM, the CPQ is zero.
If compared with the consumed tokens, \ours~ uses significantly fewer tokens because it requires fewer examples as demonstrations and does not use self-consistency.
As a result, the token cost is only about 1/10 that of other ICL-based methods.
\revise{We also provide a detailed analysis of token consumption in the Appendix \ref{appendix:consumed_token} and model detail in Appendix \ref{appendix:detail_of_CPQ}.}

\begin{table}[t]
\centering
\resizebox{\columnwidth}{!}{
    \begin{tabular}{lrrrc@{\hspace{0.1\tabcolsep}}rrr}
    \toprule
    \multirow{2}{*}{\textbf{Methods}}  & \multicolumn{3}{c}{\textbf{GrailQA}} & & \multicolumn{3}{c}{\textbf{GraphQ}}  \\    
   \cmidrule{2-4}  \cmidrule{6-8}
    & \textbf{TPQ}  & \textbf{QPQ}  & \textbf{CPQ} & & \textbf{TPQ}  & \textbf{QPQ}  & \textbf{CPQ}   \\
    \cmidrule{1-4} \cmidrule{6-8}
    KB-BINDER  & 51.2   & 3,297.7  & 0.010 & & 84.0   & 2,113.8  & 0.024  \\
    AgentBench  & 40.0  & 7.4   & 0.034  & & 65.1  & 7.2  &  0.035 \\
    QueryAgent &  16.6  &  \textbf{5.2} & 0.019 & & 15.3   & \textbf{6.2}  & 0.021  \\ 
    \cmidrule{1-4} \cmidrule{6-8}
    \ours & \textbf{4.5} & 256.8 & \textbf{0.000} & & \textbf{13.0} & 1,094.6 & \textbf{0.000} \\
     \bottomrule  
    \end{tabular}    
}
\caption{Efficiency analysis with ICL-based~(KB-Binder) and Agent-based methods~(AgentBench and QueryAgent). TPQ, QPQ, and CPQ denote the time cost (seconds), number of SPARQL query times, and open-source model invocation cost (\$) per question.} 
\label{tab:efficiency} 
\end{table}

% \subsubsection{Result with Training Dataset}
\subsubsection{Performance on different sizes of LLM}
In real-world applications, large and powerful LLMs are not always accessible or affordable. 
Therefore, we further analyze the performance of various methods across different model sizes.
% We further analyze the performance under varying model sizes. 
We compare the Agent method (QueryAgent), ICL method (KB-Binder), and the retrieval-augmented ICL method (KB-BINDER-R).
As shown in Table \ref{tab:different_llm},
our approach demonstrates remarkable adaptability from \textit{qwen-1.5B-instruct} to \textit{gpt-4o-mini}.
With just the 1.5B model, our method already surpasses the previous best-performing method, while at 7B, it only slightly lags behind the closed-source model. 
% Upon reaching the 32B model, our performance aligns with that of the closed-source models. 
The Agent method has strong generalization capabilities but is heavily reliant on the planning and self-correction abilities of the most advanced LLM, which smaller models do not excel at. 
For models below 72B, the performance of the QueryAgent is essentially unusable. 
For the 72B model, the performance of QueryAgent is still inferior to that of the ICL method using a model of the same size and ultimately failing to exceed closed-source model performance. 
Regarding the ICL methods, previous works typically experiment on the strongest closed-source models without testing their performance on open-source models. 
We demonstrate here the performance of ICL on open-source models, revealing that the latest open-source models can reach or even surpass the capabilities of the GPT series models in certain tasks. 
This provides a feasible assurance for continuing research in semantic parsing based on closed-source models.

\begin{table}[t]
\centering 

    \resizebox{\columnwidth}{!}{
    \begin{tabular}{lcccccc}
    \toprule
    & 1.5B & 7B & 32B & 72B & 4o-mini & $\Delta$\\
    \midrule
    QueryAgent    & 10.3  & 16.1 & 50.8 & 58.5 & 62.3 & 52.0 \\
    KB-Binder    & 20.2 & 39.8 & 51.3 & 50.6 & 47.0 & 31.1 \\   
    KB-Binder-R  & 27.6  & 55.0 & 59.4 & 63.8 & 58.0 & 36.3\\
    \midrule
    \ours     & \textbf{61.3}  & \textbf{65.3} & \textbf{67.4} & \textbf{67.5} & \textbf{67.7} & \textbf{9.4}\\
    \bottomrule 
    \end{tabular}
}
\caption{The performance on GrailQA with different base model sizes. The 1.5B, 7B, 32B, and 72B represent the Qwen 2.5 instruct models family, while 4o-mini indicates GPT-4o-mini. We experiment on 500 random sampling questions. $\Delta$ indicates the max performance gap between different models.}
\label{tab:different_llm}

\end{table}

\begin{table}
    \centering 
    
    \resizebox{\columnwidth}{!}{
    \begin{tabular}{lccc}
    \toprule
   & \textbf{GrailQA}  & \textbf{GraphQ} &  \textbf{KBQA-Agent} \\
   \midrule
    \ours          & \textbf{69.0} & \textbf{50.6} & \textbf{46.5} \\ 
    ~~~~w/o Query Textification & 64.9          & 46.6          & 39.6  \\
    ~~~~w/o Re-Ranking          & 59.9          & 38.7          & 19.3 \\
    ~~~~w/o Synthetic Question  & 67.3          & 49.5          & 45.5 \\
  \bottomrule
    \end{tabular}
    } 
\caption{Ablation study of each component on GrailQA.}
    \label{tab:ablation}    
\end{table}

\subsubsection{Ablation Study}

Table~\ref{tab:ablation} presents the impact of distinct components on model performance across three datasets. Compared to the full model, removing the query textification component leads to a noticeable drop, particularly on KBQA-Agent (-6.9), highlighting the importance of bridging the gap between the text embedding model and logic form.
% indicating its significant contribution to performance. 
The removal of the re-ranking component results in the largest performance decrease, with reductions of 9.1, 11.9, and 27.2 on GrailQA, GraphQ, and KBQA-Agent, respectively, underscoring the importance of the re-ranking step. 
In contrast, excluding synthetic question generation yields more modest declines, suggesting it is less critical than the other components but still beneficial for KBQA-Agent. 
It is unexpected, but from another aspect, it indicates that even only using the synthetic query as the demonstration the performance is also competitive.

\subsubsection{Transferability to Text2SQL}
We adapted our framework to the Text2SQL task to demonstrate the generality of our approach in other semantic parsing tasks. 
Employing the WikiSQL dataset, we compare \ours~ with both the fine-tuned and non-fine-tuned methods.
Among them, StructGPT and Readi are 32-shot and 7-shot methods, respectively.
AgentBench and QueryAgent both use 1 shot.
As shown in Table \ref{tab:wikisql}, with merely 10 synthetic examples as demonstrations, our method surpasses prior methods with 32 manually annotated examples and also outperforms the best 1-shot method, all while incurring a lower cost and smaller model. Besides, \ours~can even surpass a fine-tuned method with 3B model.

\begin{table}[t]
\centering

\resizebox{0.95\columnwidth}{!}{
    \begin{tabular}{lrr}
     \toprule 
    \textbf{Methods} & \textbf{Acc.}\\
    \midrule
    % \textbf{\textit{Full dataset}} \\
    % finetune来自din-sql的结果
    RESDSQL-3B + NatSQL*~\cite{li2022resdsql}  & 79.9 \\
    T5-3B+PICARD*~\cite{scholak2021picard}  & 75.1 \\
    \midrule 
    % \textbf{\textit{Dozens of annotations}} & \\
    % Davinci-003  & 49.1  & \\    
    % ChatGPT  & 51.6  & \\ 
    % StructGPT(Davinci-003)   & 64.6 & \\    
    StructGPT (ChatGPT)~\cite{jiang2023structgpt} & 65.6 & \\   
    Readi~\cite{cheng2024call} &  66.2 & \\
    % \midrule 
    % \textbf{\textit{One annotation}} \\
    AgentBench~\cite{liu2023agentbench} &  57.6 & \\
    QueryAgent~\cite{huang2024queryagent} &  72.5 & \\
    % \textbf{\textit{Zero annotation}} & \\
    \midrule
    \ours & \textbf{75.5}  &  \\ 
     \bottomrule 
    \end{tabular} 
}
\caption{Results on WikiSQL. * indicates fine-tuned. 
% Other baselines are few-shot or one-shot methods 
% Our method is 10-shot while other few-shot methods are 32-shot.
}
\label{tab:wikisql}
\end{table}

\section{Conclusion}
In this paper, we explore two critical challenges in the semantic parsing task: reliance on annotated data and poor generalization on non-I.I.D. cases.
We proposed a novel method called \ours, which automatically synthesizes examples that are most relevant to the test data and utilizes them as demonstrations for in-context learning.
Remarkably, \ours achieves the best performance among all non-fine-tuned methods across three complex KBQA datasets and one Text2SQL dataset, especially on GrailQA and KBQA-Agent~(7.7 and 12.2 F1 points, respectively).
While achieving impressive performance, \ours~also exhibits the following practical properties:
1) It does not require any annotated data.
2) It is effective even with a model size of just 7B parameters.
3) The synthetic data is generated online.
4) It exhibits superior generalization, robustness, and speed.
This work highlights the potential of leveraging synthetic data in semantic parsing, and we hope that \ours~can serve as a valuable foundation for developing more practical systems in this field.

\section*{Limitations}
We would like to discuss some limitations of our work. 
First, in this paper, we validate \ours~on two specific semantic parsing tasks: KBQA and Text2SQL. 
While these tasks demonstrate the potential of our approach, further exploration across a broader range of tasks that involve transforming natural language into logical forms could strengthen the generalizability of \ours. 
Additionally, we have not yet investigated the feasibility of our synthetic data generation method in other paradigms, such as agent-based or fine-tuned models.
We would like to adapt \ours~to these paradigms in future work.

% \bibliography{custom}

% \input{acl_latex.bbl}
\bibliographystyle{IEEEtran}
\bibliography{acl_latex.bbl}
\newpage
\clearpage

\appendix

\section{Preliminary}

\subsection{Knowledge Base Question Answering}
We introduce related concepts and the task of knowledge base question answering as follows. Let~$E$ be a set of entities, $P$~be a set of relations, $C$~be a set of classes, and $I$~be a set of literals. A \emph{knowledge base} $\mathcal{K} \subseteq E \times P \times (E \cup C \cup I)$ is a set of triples~$(s, p, o)$, where~$s \in E$ is a subject entity, $p \in P$ is a predicate, and~$o \in (E \cup C \cup I)$ is an object entity, class or literal value. 

The task of knowledge base question answering can be formalized as learning a function~$f$ that takes a natural language question~$q$ as input, and outputs a structured query~$q' = f(q)$ in a formal language such as SPARQL~\cite{sparql}. The structured query~$q'$ should ideally encode the entities, relations, and constraints specified by the input question~$q$, such that executing~$q'$ over the knowledge base~$\mathcal{K}$ yields the correct answer of~$q$. 

\subsection{In-Context Learning}
In-context learning~\cite{brown2020language} allows LLMs to perform new tasks by simply observing examples provided within the input, without updating their internal parameters. Intuitively, the model ``learns'' from the context and uses it to generate appropriate responses for similar tasks, relying on patterns it recognizes from the given examples.

More precisely, let $[x_1, x_2, \cdots, x_n]$ be a sequence of input tokens representing natural language texts, and let $[y_1, y_2, \cdots, y_n]$ be a corresponding sequence of output tokens representing the desired task, which in our context are structured queries. A LLM, denoted as~$f_\theta$, is a function parameterized by~$\theta$, which takes an input sequence and predicts the next token or sequence of tokens. In-context learning refers to the ability of a pretrained LLM to learn and adapt to a specific task purely by conditioning on a sequence of examples $S = [(x_1, y_1), (x_2, y_2), \cdots, (x_n, y_n)]$, provided as part of the input context, without updating the model parameter~$\theta$. In contrast to traditional learning paradigms that require parameter updates via gradient descent, the LLM uses the provided examples to infer the underlying task and generate predictions for a new input~$x_{n+1}$. 

\section{Datasets}
\label{appendix:datasets}
We conduct experiments on four KBQA datasets and one Text2SQL dataset, their statistics are shown in Table~\ref{tab:dataset_static}.
\begin{itemize}
    \item \textbf{GrailQA}~\cite{gu2021beyond} is one of the most popular complex KBQA datasets. It divides the dataset into three levels of generalization, \textit{i.e.,} I.I.D, Compositional, and Zero-shot.
    
    \item \textbf{GraphQ}~\cite{su2016graphquestions} is a challenging dataset that only consists of Compositional questions.
    
    \item \textbf{KBQA-Agent}~\cite{gu2024middleware} is a mixed dataset of the most difficult questions from four datasets (\textit{i.e.,} GrailQA~\cite{gu2021beyond}, ComplexWebQuesiton~\cite{talmor2018the}, GraphQ~\cite{su2016graphquestions}, and WebQSP~\cite{yih2016the}).

    \item \textbf{MetaQA-3Hop}~\cite{zhang2017variational} is the most difficult 3-hop subset of a large-scale KBQA dataset based on Wiki-Movies KG. 
    
    \item \textbf{WikiSQL}~\cite{zhong2017Seq2SQL} is a large-scale complex Text2SQL dataset which requiring comparison, aggregation and arithmetic operations.
    
\end{itemize}

\begin{table}[t]
\centering
\resizebox{0.9\columnwidth}{!}{
\begin{tabular}{crrr}
    \toprule
    \textbf{Datasets} & \textbf{Train}  & \textbf{Dev} & \textbf{Test}  \\
    \midrule          
    \textsc{GrailQA} & 44,337  & 6,763  & 13,231 \\
    \textsc{GraphQ} & 2,381  &  -   &  2,395 \\
    \textsc{KBQA-Agent} & -  &  -   &  500 \\
    \textsc{MetaQA-3hop} & 114,196 &  14,274 &  14,274 \\
    \textsc{WikiSQL} & 56,355  &  8,421   &  15,878 \\ 
     \bottomrule   
\end{tabular} 
}
\caption{Statistics of datasets.} 
\label{tab:dataset_static}
\end{table}

\section{Baselines}
\label{appendix:baselines}
\subsection{Seq2Seq Fine-tuning Methods}

\begin{itemize}
    
    \item \textbf{ArcaneQA}~\cite{gu2022arcaneqa} is a generation-based method that incrementally synthesizes a program by dynamically predicting a sequence of subprograms. It prunes the search space by constrained decoding.
    
    \item \textbf{Pangu}~\cite{gu2023dont} leverages the discrimination ability of language models to build queries in an incremental manner. It consists of a symbolic agent to collect valid candidate plans and an LM to select the most likely one. 

\end{itemize}

\subsection{ICL-based Method}

\begin{itemize}
    \item \textbf{Pangu-ICL}~\cite{gu2023dont} is a ICL version of Pangu. The result is based on output distribution of the LLMs, which require access to model parameters.

    \item \textbf{KB-Binder}~\cite{li2023few} propose an ICL-based method for few-shot KBQA by feeding LLM with some (question, S-expression) pairs. 

    \item \textbf{KB-Coder}~\cite{nie2023codestyle} further optimizes KB-Binder by changing the target format from S-expression to code-style logic form.

    \item \textbf{Readi}~\cite{cheng2024call} propose a Reasoning-Path-Editing framework that initially generates a reasoning path given a query, then instantiates the path. Editing is triggered only when necessary. 

    \item \textbf{BYOKG}~\cite{agarwal2024bring} does not need manually annotated dataset. It first explores the KB to collect synthetic datasets within a day~(\textit{e.g.,} 10K data in 10 hours), then uses the synthetic dataset for bottom-up reasoning.
\end{itemize}

\redchapter{
\subsection{Agent Fine-tuning Methods}
\begin{itemize}
    \item \textbf{KG-Agent}~\cite{jiang2024kgagent} enables a small LLM to actively make decisions until finishing the reasoning process over KGs through fine-tuning an Agent.
    \item \textbf{DARA}~\cite{fang2024dara} propose a Decomposition-Alignment-Reasoning Autonomous Language Agent which can be efficiently trained with a small number of high-quality reasoning trajectories.
\end{itemize}
}
\subsection{Agent-based Method~(Training-free)}

\begin{itemize}
    \item \textbf{AgentBench}~\cite{liu2023agentbench} model KBQA as a tool learning task and outfitting LLM with an array of KG-querying tools such as ``get\_ relation'', ``argmax'' and ``intersection''.

    \item \textbf{FUXI}~\cite{gu2024middleware} design customized tools acting as middleware between LLMs and complex environments. They also incorporate decoupled generation and error feedback to boost performance. 

    \item \textbf{QueryAgent}~\cite{huang2024queryagent} step-by-step build the target query and use an environmental feed-based self-correction method to reduce hallucination.
\end{itemize}

\section{Levels of Generalization}
\label{appendix:generalization_level}
In the context of KBQA, the three levels of generalization---I.I.D. generalization, compositional generalization, and zero-shot generalization---refer to the capability of models to handle increasingly challenging and diverse types of questions~\cite{gu2021beyond}. 
In particular, I.I.D. generalization refers to the model’s ability to correctly answer questions that are sampled from the same distribution as the training set, which assumes that the test data follow similar patterns and schema. On top of that, compositional generalization refers to the model’s ability to handle novel combinations of KB items (e.g., entities and relations) that were seen during training, but in configurations that the model has not encountered before. Finally, zero-shot generalization refers to the model’s ability to answer questions involving entirely new KB items, such as unseen entities and relations that were never presented in the training set.

\citet{gu2021beyond} argue that relying solely on datasets under I.I.D. settings limits the practical use of KBQA models, as real-world questions often involve unfamiliar entities or require novel reasoning. 
This limitation is evident in the degraded performance of existing KBQA methods under compositional and zero-shot settings compared to I.I.D. settings.
Therefore, practical KBQA models should be equipped with built-in generalization capabilities across all three levels to better handle diverse, real-world questions.

\section{More Details}
\subsection{Other Experiment Settings}
\ours and Pangu used 10 demonstrations across all datasets. KB-Binder and KB-Coder used 40 shots for GrailQA, 100 shots for GraphQ, and 5 shots for MetaQA, with KB-Binder employing 20 shots for KBQA-Agent. AgentBench, FUXI, and QueryAgent are all 1-shot methods. The experiments using models with 1.5B, 7B, 32B, and 72B parameters were conducted on 1, 1, 4, and 8 A100 GPUs, respectively.

\subsection{Entity Linking Detail}
\label{subsec:entity_linking_detail}

\revise{
The detailed entity linking results of all compared methods are listed in Table~\ref{tab:entity_linking}.
For KBQA-Agent and MetaQA-3Hop, all compared methods use golden linking result.
For GrailQA and GraphQ, most compared methods use the entity linking result by Pangu~\cite{gu2023dont}.
Therefore, we follow their setting to make a fair comparison.
}

\begin{table*}[ht]
    \centering
    
    \begin{tabular}{ccccc}
    \toprule
  \textbf{Method}     & \textbf{GrailQA}	& \textbf{GraphQ} &	\textbf{KBQA-Agent}&	\textbf{MetaQA-3Hop} \\
       \midrule
KB-Binder & 	FACC1+BM25	 & FACC1+BM25 & 	Golden &	Exact match\\
KB-Coder & 	FACC1+SimCSE	 & FACC1+SimCSE	 & - & 	-\\
Pangu(ICL)	 & Pangu & 	Pangu	 & Golden & 	-\\
FUXI	 & - & 	- & 	Golden	 & -\\
AgentBench & 	Pangu & 	Pangu	 & Golden	 & -\\
QueryAgent & 	Pangu & 	Pangu & 	- & 	Exact match\\
\midrule
\textbf{TARGA} & 	Pangu & 	Pangu	 & Golden	 & Exact match\\
         \bottomrule
    \end{tabular}
\caption{The detailed source of entity linking data for compared method.
    The original performance of AgentBench is based on the Golden entity linking result. For a fair comparison, we report the perfomance with the Pangu entity linking result re-implemented by QueryAgent~\cite{huang2024queryagent} in Table \ref{tab:main_result}.}    
    \label{tab:entity_linking}    
\end{table*}

\revise{
Among all the methods compared, KB-Binder and KB-Coder chose a pipeline that was different from the other methods.
Other methods first get the entity linking results and then generate logic forms based on them.
KB-Binder and KB-Coder first generate a logic form draft and then bind each KB-item in the logic form draft to the knowledge base.
For KB-Binder and KB-Coder, the elements requiring linking depend on the generated draft, making it less suitable for directly using the entity linking results as other methods. 
}

\subsection{Logic Form Design}
\label{appendix:lf_design}
Inspired by PyQL~\cite{huang2023markqa}, we designed a simplified logical form for constructing queries, which uses a series of functions for semantic expression, making it easier for LLMs to learn and enabling seamless translation into SPARQL.
Table~\ref{tab:pyql} shows all functions that we used.

\redchapter{
\subsection{Query Textification Detail}
\label{appendix:textification}
To improve the performance of the text embedding model, we transform the synthesized query into a format closer to natural language by implementing a simple yet effective rule-based parsing program. 
An example of query textification is provided in Table~\ref{tab:textification}. 

Specifically, we first replace the entities and relations in the triples with their labels, concatenating the subject, predicate, and object to create a description.
For two nested descriptions (multi-hop), 
we represent them in a hierarchical form using ``a have b.''.
For two conjunction descriptions~(multi-constraint), we connect the two parts with ``and.''
We also applied some simple processing to the filters, such as using "more than" and "less than" to connect comparison objects.

\subsection{Statistics of Query Construction}
\label{appendix:statics_of_query_construction}
As shown in Table \ref{tab:query_construct}, we present the statistics of the synthetic query in Section \ref{subsec:syhthetic_query_generation}.

\begin{table}[ht]
\centering

\resizebox{0.95\columnwidth}{!}{
\begin{tabular}{lrrrr}
\toprule
\textbf{Datasets} & \textbf{Size} & \textbf{Cvg.} & \textbf{TPQ} & \textbf{QPQ} \\
\midrule
GrailQA & 17.4 & 0.79 & 2.99 & 256.8 \\
GraphQ & 89.3 & 0.69 & 11.41 & 1,094.6 \\
KBQA-Agent & 173.1 & 0.78 & 20.07 & 2,270.3 \\
MetaQA-3Hop & 25.9 & 1.00 & 0.05 & 56.1 \\
WikiSQL & 18.1 & 0.75 &  0.16 & 88.6 \\
\bottomrule
\end{tabular}
}
\caption{Statistics of query construction. 
Size represents the average number of valid queries per question.
Cvg. refers to coverage, indicating the proportion of questions with at least one correct~(F1 = 1) in synthetic queries. 
TPQ and QPQ denote the running time (in seconds) and the number of query attempts per question, respectively.}
\label{tab:query_construct}
\end{table}

\subsection{Natural Language Question Generation Detail}
\label{appendix:nlq_generation_detail}

We directly use the pseudo-questions generated by query textification for building (query, question) pairs as demonstrations in QA phase, and the performance is good enough.
We also experimented with using an LLM (GPT or Qwen) to generate questions during this phase, but the downstream QA performance remained nearly unchanged, with a difference of less than 0.5 F1. Moreover, leveraging an LLM to generate corresponding questions requires labeling a few examples as demonstrations and incurs an extra time cost (less than 1 second).

In fact, even without generating questions and using only the queries as demonstrations, the performance showed only a slight decline (see the ablation study in Table 6). We recommend choosing between using rules or an LLM based on specific scenarios:
\begin{itemize}
    \item If you want to use TARGA for ICL-style QA, the quality of question generation at this stage does not need to be particularly high. Heuristic rules are sufficient enough.
    % Even without generating natural language questions can still achieve satisfactory performance.
    \item If you want to construct standard parallel corpora for fine-tuning or other purposes, 
    we recommend you use an LLM along with some demonstrations to generate higher-quality natural language questions.
\end{itemize}

\subsection{Prompt for Question Answering}

We provide example prompts for \ours on the KBQA and Text2SQL tasks In Tables \ref{tab:show_prompt} and \ref{tab:show_prompt2}.
It consists of some synthetic demonstrations~(NLQ-Query pairs) and the test question.
For the KBQA task, we provide the entity linking results.
For the Text2SQL task, we provide the headers of the table.

\begin{table*}

\resizebox{0.98\textwidth}{!}{
    \begin{tabular}{p{3cm} p{11.5cm}}
    \toprule
    \textbf{Functions} & \textbf{Brief Description} \\ 
    \midrule
    triplet(s, p, o) & Add a condition that the subject s is linked to the object o via a relation p.\\
    argmax(v) & Add a condition that the variable v must have the maximum value.\\
    argmin(v) & Add a condition that the variable v must have the minimum value.\\
    filter(v, op, value) & Add a condition that the variable v must meet, where the operator op is one of [<, >, <=, >=].\\
    type(v, t) & Specify that the variable v belongs to the class~(entity type) t.\\
    count(v) & Set the count of variable v as the final answer.\\
    answer(v) & Set the variable v as the final answer.\\
    \bottomrule   
    \end{tabular}
}
\caption{Functions used in our logic form.}
\label{tab:pyql}
\end{table*}

\section{Further Analysis}
\subsection{The performance difference between Compositional and zero-shot setting}
In Table \ref{tab:generalization}, we notice that the performance of the compositional setting is lower than the zero-shot setting for \ours, but this trend is reversed for other compared methods.
Our perspective is outlined as follows:
\begin{itemize}
    \item This is a matter of relative performance:
If the performance of TARGA on zero-shot performance is above the normal level, then compositional will appear lower than zero-shot, and vice versa for comparative methods.
If the performance of other methods on zero-shot setting is very poor, the performance of compositional setting will appear relatively high. 
    \item Setting aside whether the test questions are seen in the training set or not, there are significant inherent difficulty differences among the three generalization levels of GrailQA: compositional is the hardest, followed by i.i.d., and zero-shot is the easiest.
\end{itemize}

Firstly, we explain why other methods exhibit lower zero-shot performance:

The zero-shot setting implies that the given question contains entities or relationships not seen in the training set.
The compositional setting involves combinations of entities and relationships that have not been seen together, though each individual element is exist in the training set.
Note that the comparative methods we used are, to some degree, influenced by the training set.
Retrieving several examples from the training set can still essentially cover all the elements needed to solve a question, but this is not the case with zero-shot. This may explain why compared methods exhibit lower zero-shot performance than compositional setting.
Besides, for GraphQ, the 100\% Compositional dataset, \ours~exhibit more significant improvement compared to KB-Binder and KB-Coder than GrailQA (25\% Compositional), indicating that \ours~does not harm the performance of the compositional setting.

Secondly, we explain why our compositional performance is lower than zero-shot:

We believe that the fundamental reason for this phenomenon lies in how the dataset is split and the inherent difficulty of the problems themselves.
For TARGA, there is no substantial distinction between input questions, as each one is treated as an unseen question.
However, the generalization level of a question is determined solely by whether it appears in the training set, irrespective of the question’s inherent difficulty.
As a result, TARGA serves as a fair model for evaluating a question’s inherent difficulty, as it is unaffected by any static dataset distribution.
We hypothesize that TARGA’s performance across the three generalization levels reflects significant differences in difficulty within the GrailQA dataset, with the compositional setting being the most challenging, followed by i.i.d., and zero-shot being the easiest.

To support this hypothesis, Table~\ref{tab:grailqa_difficulty} reports the average length of the final SPARQL queries for questions across the three generalization levels. In general, longer SPARQL queries correspond to more complex questions (i.e., higher difficulty). Compositional questions yield the longest queries and the lowest performance, whereas zero-shot questions are the simplest and achieve the highest performance. These statistical results strongly validate our earlier conjecture.

\begin{table}[t]
    \centering
    
    \resizebox{0.48\textwidth}{!}{
    \begin{tabular}{lcc}
    \toprule
    \textbf{Generalization level} & \textbf{Difficulty} & \textbf{Performance}	\\
    \midrule
    Compositional & 	5.1 & 	62.2\\
    I.I.D. & 	4.8	& 68.4 \\
    Zero-shot & 	4.5	& 71.7 \\                
    \bottomrule
    \end{tabular}
    }
\caption{Difficulty and performance of TARGA across three generalization settings on GrailQA. The difficulty is measured by the average length of SPARQL queries, and the performance shows TARGA's F1 score.}
    \label{tab:grailqa_difficulty}
\end{table}

\subsection{Further Analysis on Robustness}
\label{appendix:robust_entity}
In Section \ref{subsec:robustness_analyze}, we have analyzed the effect of randomly replacing relations in demonstrations.
Here, we present the performance of randomly replacing entities in the demonstrations.
Similarly to \ref{subsec:robustness_analyze}, we corrupt N demonstrations by replacing one entity with another randomly selected entity from the training set (with a candidate pool of 32K entities).
The results in Figure~\ref{fig:attack_ent} indicate that replacing the entity has a minimal impact across all three settings.

\begin{figure}[t]
    \centering
    \includegraphics[width=0.9\linewidth]{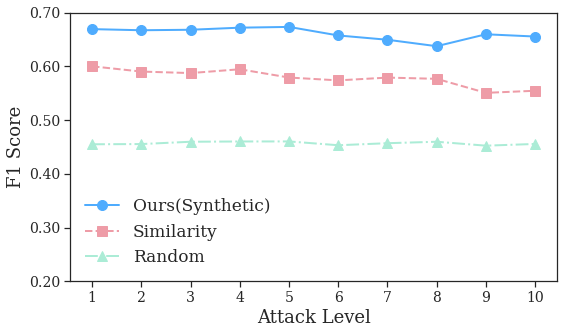}
    \caption{Performance under attack setting(entity) on 1,000 randomly sampled GrailQA questions.}
    \label{fig:attack_ent}
\end{figure}

\begin{table*}

\begin{tabular}{p{15cm}}
    \toprule 
    \textbf{Queries with textification results} \\ 
    \midrule
    \textbf{Query~1:} \\
    triplet(?v0, measurement\_unit.mass\_unit.weightmass\_in\_kilograms, ?v1) \\
    argmin(?v1)\\
    answer(?v0)\\
    \textbf{Textification Result:} \\ what mass\_unit, mass\_unit has weightmass\_in\_kilograms, when weightmass\_in\_kilograms is the smallest\\
       \midrule
    \textbf{Query 2:} \\
    triplet(?v0, boats.ship\_class.date\_designed, ?v1) \\
    argmax(?v1) \\
    answer(?v0)\\
    \textbf{Textification Result:} \\ 
    what ship\_class, ship\_class has date\_designed, when date\_designed is the largest\\
    \midrule
    \textbf{Query 3:} \\
    triplet(?v0, spaceflight.rocket\_engine.designed\_by, [rocketdyne]) \\
    triplet(?v0, spaceflight.rocket\_engine.isp\_sea\_level, ?v1) \\
    filter(?v1, <=, 260.0)\\
    answer(?v0) \\
   \textbf{Textification Result:} \\ 
   what rocket\_engine, rocket\_engine has rocketdyne, rocket\_engine has isp\_sea\_level, when isp\_sea\_level no more than 260.0\\ 
    \bottomrule   
\end{tabular}
\caption{Examples of Query Textification.}
\label{tab:textification}
\end{table*}

\subsection{Further analysis on ranking}

Considering that a poor ranking might incorrectly identify the top 1 or top 2 results, while the top 10 overall might still be generally correct. Therefore, we conducted a more convincing experiment using only the top 1 data as the demonstraion to better illustrate the impact of query textification on ranking quality.

The results are shown in Table~\ref{tab:more_ranking_abalation}: we found that in this setting, the decline without query textification was generally more pronounced (9.9, 7.6, and 3.1 on three datasets) than Table \ref{tab:ablation} (4.1, 4.0 and 6.9 on three datasets), indicating that query textification can enhance ranking quality, thereby further improving QA performance.

\begin{table}[ht]
    \centering
    
    \begin{tabular}{lrrr}
    \toprule
         & \textbf{GrailQA}	&  \textbf{GraphQ}	&  \textbf{KBQA-Agent} \\
         \midrule
        w/ QT	&  61.46	&  44.31&  	37.53\\
        w/o QT	&  52.55	&  36.76	&  34.48\\
\bottomrule
    \end{tabular}
\caption{Abalation study of Query Textification~(QT) when only using Top 1 candidate as the demonstration~(experiment on 500 randomly selected examples).}
    \label{tab:more_ranking_abalation}
\end{table}

\subsection{Runtime Breakdown}
\label{appendix:runtime_breakdown}
We present a detailed runtime breakdown to provide a clearer understanding of the time costs associated with each stage, using GrailQA as an example.

\textbf{(1). Entity linking: nearly negligible~(0 s)}:
We use the cache provided by previous work for fair comparison. Most previous non-finetuning methods either rely on cached results or the golden linking results. Details are provided in Section \ref{subsec:entity_linking_detail}.

\textbf{(2). Relation linking: very fast~(0.60 s)}:
Relation linking consists of three steps:

a). Obtain the relation embedding (0 s, cached in advance). Note that the relations in Freebase are fixed and limited, so they only need to be cached once in advance.

b). Obtain the question embedding~(0.52 s). 

c). Calculate the embedding similarity between question embedding and all relation embeddings (0.08s).
We utilized FAISS~\cite{johnson2019billion} to accelerate similarity computations.

\textbf{(3). Query construction: the slowest part} \textbf{(2.99 s for GrailQA)}:

This is the most time-consuming stage, which is described in detail in Section \ref{subsec:syhthetic_query_generation}.

\textbf{(4). Reranking: very fast~(0.11 s)}:
This is implemented by bge-reranker.
Since we carefully controlled the number of candidate queries during the construction process, the final set for ranking is relatively small.
Details about the size of the candidate queries are provided in Table \ref{tab:query_construct}.
For instance, only an average of 17.4 candidate queries were involved in the reranking stage, which contributed to its speed.

\textbf{(5). Question generation: nearly negligible} \textbf{(0 s)}:
Since we use the pseudo-questions generated by query textification~(implemented by simple rules) as the corresponding question of a candidate query, this time cost is negligible.

\textbf{(6). In-Context Learning QA~(0.78 s)}:
This is a simple ICL request with about 700 input tokens; a normal inference speed is generally less than 1 second.
The time difference between using locally deployed open-source models (Qwen) and online closed-source models (GPT) is not significant.
For closed-source models (GPT), a normal request speed is generally on the order of seconds.
For open-source models, the specific speed depends on your hardware configuration. We deployed a 7B model using a single 80GB A100 GPU and a 72B model using eight A100 GPUs, leveraging vLLM~\cite{kwon2023efficient} for deployment. Using more devices can further speed up.

\subsection{Clarifications on the Computational Details of CPQ}
\label{appendix:detail_of_CPQ}
We follow the OpenAI official price calculation method~\footnote{\url{https://openai.com/api/pricing/}} as reported in QueryAgent~\cite{huang2024queryagent} to calculate CPQ in Table \ref{tab:efficiency}.
\ours~only relying on an open-source model, except for Section \ref{subsec:kb_items_retrieval} use \textit{ada-v2} for obtain the embedding and question.
However, this part of the cost is less than 1e-5~\$ per question, therefore, the CPQ is 0~\$ when retaining three decimal places in Table \ref{tab:efficiency}.

The cost in Section \ref{subsec:kb_items_retrieval} consist of two parts: relation embedding and question embedding.
The cost of caching embeddings for all relations in Freebase (20K relations) is less than 0.01~\$, and this cost is one-time; once cached, it does not increase regardless of the number of questions solved. 
For obtaining the embeddings for the test question itself, the average cost per question is less than 1e-5~\$, which rounds to approximately 0.00~\$ when retaining three decimal places in Table \ref{tab:efficiency}.  
In fact, the cost of obtaining embeddings for all questions across the three Freebase datasets~(comprising 10,758 questions) does not exceed 0.05~\$.
In a word, the cost for using open-source model  can
In summary, the close-source model invocations cost of \ours~is negligible.

\subsection{Analysis of Token Consumption}
\label{appendix:consumed_token}
Since \ours~basically does not rely on closed-source models,
to make the inference cost more comparable,
we additionally provide a comparison of the total tokens consumed (input tokens + output tokens) for a more comprehensive evaluation.
The result is shown in Table~\ref{tab:consumed_tokens}.
Note that the substantial difference in token consumption by KB-Binder across the two datasets primarily stems from the use of more demonstration: 100 shots for GraphQ compared to 40 shots for GrailQA.

\begin{table}[ht]
    \centering
   
    \begin{tabular}{lrr}
    \toprule
    &  \textbf{GrailQA}	&  \textbf{GraphQ}\\
    \midrule
KB-Binder	&  6,138 + 189	&  15, 166 + 206\\
AgentBench	&  19, 783 + 934	&  20, 143 + 968\\
QueryAgent	&  12, 277 + 391&  	13, 453 + 391\\
\midrule
\textbf{TARGA}	& \textbf{638} + \textbf{36}&  	\textbf{734} + \textbf{38}\\    
\bottomrule
    \end{tabular}
\caption{Average consumed tokens (input + output).}
    \label{tab:consumed_tokens} 
\end{table}

\subsection{Case Study}
\label{appendix:case_study}
Table~\ref{tab:case_study} presents a detailed demonstration of how \ours operates with a new question.

\onecolumn
% \resizebox{\textwidth}{!}{
\begin{longtable}{@{}p{3.5cm}p{12cm}@{}}  % 调整列宽
\toprule
\textbf{Pipeline} & \textbf{Details} \\ 
\midrule
\endfirsthead

\toprule
\textbf{Pipeline} & \textbf{Details} \\ 
\midrule
\endhead
\textbf{Question} & the camera with a sensor from canon and a compression format of jpeg (exif 2.21) uses which viewfinder? \\
\midrule
\textbf{Candidate Entities} & \{canon: m.01bvx1, jpeg (exif 2.21): m.03h4lt3 \} \\
\midrule
\textbf{Candidates Relations} & [\underline{digicams.digital\_camera.viewfinder\_type}, \newline \underline{digicams.camera\_compressed\_format.cameras}, \newline
digicams.digital\_camera.sensor\_type, \newline
digicams.camera\_uncompressed\_format.cameras, \newline
digicams.camera\_iso.cameras, \newline
digicams.camera\_format.cameras, \newline
\underline{digicams.camera\_sensor\_manufacturer.cameras}, \newline
...,\newline
digicams.camera\_white\_balance.cameras] \\
\midrule
\textbf{Query Construction} & \textbf{Query~1}\newline 
\textit{triplet([jpeg (exif 2.21)], digicams.camera\_compressed\_format.cameras, ?v0) \newline 
answer(?v0)} \newline
\textbf{Query~2}\newline 
\textit{triplet([jpeg (exif 2.21)], digicams.camera\_compressed\_format.cameras, ?v0) \newline 
triplet(?v0, digicams.digital\_camera.viewfinder\_type, ?v1)\newline 
answer(?v1)} \newline
...... \newline
\textbf{Query~56} \newline 
\textit{triplet([jpeg (exif 2.21)], 
digicams.camera\_compressed\_format.cameras, 
?v0) \newline 
triplet(?v0, 
digicams.digital\_camera.viewfinder\_type, 
?v1) \newline 
triplet([canon], 
digicams.camera\_sensor\_manufacturer.cameras, 
?v0) \newline 
answer(?v1)} \newline
...... \newline
\textbf{Query~71} \newline 
\textit{triplet([canon], 
digicams.digital\_camera\_manufacturer.cameras, 
?v0) \newline 
triplet(?v0, 
digicams.camera\_storage\_type.compatible\_cameras, 
?v1) \newline 
answer(?v1)} \\
\midrule
\textbf{Query Re-ranking} & \textbf{Rank~1} \newline
\textit{triplet([jpeg (exif 2.21)], 
digicams.camera\_compressed\_format.cameras, 
?v0) \newline 
triplet(?v0, 
digicams.digital\_camera.viewfinder\_type, 
?v1) \newline 
triplet([canon], 
digicams.camera\_sensor\_manufacturer.cameras, 
?v0) \newline 
answer(?v1)} \newline
\textbf{Rank~2} \newline
\textit{triplet([jpeg (exif 2.21)], 
digicams.camera\_compressed\_format.cameras, 
?v0) \newline 
triplet(?v0, 
digicams.digital\_camera.viewfinder\_type, 
?v1) \newline 
triplet([canon], 
digicams.digital\_camera\_manufacturer.cameras, 
?v0) \newline 
answer(?v1)} \newline
......\newline
\textbf{Rank~71} \newline
\textit{triplet([canon], 
digicams.digital\_camera\_manufacturer.cameras, 
?v0) \newline 
triplet(?v0, 
digicams.camera\_storage\_type.compatible\_cameras, 
?v1) \newline 
answer(?v1)} \\
\midrule
\textbf{Question Generation} & \textbf{Query~1} \newline
\textit{triplet([jpeg (exif 2.21)], 
digicams.camera\_compressed\_format.cameras, 
?v0) \newline 
triplet(?v0, 
digicams.digital\_camera.viewfinder\_type, 
?v1) \newline 
triplet([canon], 
digicams.camera\_sensor\_manufacturer.cameras, 
?v0) \newline 
answer(?v1)} \newline
\textbf{Generated Question} \newline
% what is the viewfinder type of the digital camera that uses the jpeg (exif 2.21) compressed format and is manufactured by canon? 
what viewfinder\_type, jpeg ( exif 2.21 ) has cameras, cameras has viewfinder\_type, canon has cameras  \newline

\textbf{Query~2} \newline
\textit{triplet([jpeg (exif 2.21)], 
digicams.camera\_compressed\_format.cameras, 
?v0) \newline 
triplet(?v0, 
digicams.digital\_camera.viewfinder\_type, 
?v1) \newline 
triplet([canon], 
digicams.digital\_camera\_manufacturer.cameras, 
?v0) \newline 
answer(?v1)} \newline
\textbf{Generated Question} \newline
% what is the viewfinder type of the canon camera with jpeg (exif 2.21) format?
what viewfinder\_type, jpeg ( exif 2.21 ) has cameras, cameras has viewfinder\_type, canon has cameras  \newline
......\newline

\textbf{Query~10} \newline
\textit{triplet([jpeg (exif 2.21)], 
digicams.camera\_compressed\_format.cameras, 
?v0) \newline 
triplet(?v0, 
digicams.digital\_camera.sensor\_resolution, 
?v1) \newline 
triplet([canon], 
digicams.digital\_camera\_manufacturer.cameras, 
?v0) \newline 
answer(?v1)} \newline
\textbf{Generated Question} \newline
% what is the sensor resolution of the canon digital camera with jpeg (exif 2.21) format? 
what sensor\_resolution, jpeg ( exif 2.21 ) has cameras, cameras has sensor\_resolution, canon has cameras  \\
\midrule

\textbf{QA Prompt} & 
You are a powerful model for generating PyQL queries to answer natural language questions.
Here are some exemplars: \newline
\#\#\#Question \newline
what viewfinder\_type, jpeg ( exif 2.21 ) has cameras, cameras has viewfinder\_type, canon has cameras \newline
\#\#\#PyQL \newline
triplet([jpeg ( exif 2.21 )], digicams.camera\_compressed\_format.cameras, ?v0) \newline
triplet(?v0, digicams.digital\_camera.viewfinder\_type, ?v1) \newline
triplet([canon], digicams.camera\_sensor\_manufacturer.cameras, ?v0) \newline
answer(?v1) \newline

\#\#\#Question \newline
what viewfinder\_type, jpeg ( exif 2.21 ) has cameras, cameras has viewfinder\_type, canon has cameras \newline
\#\#\#PyQL \newline
triplet([jpeg ( exif 2.21 )], digicams.camera\_compressed\_format.cameras, ?v0) \newline
triplet(?v0, digicams.digital\_camera.viewfinder\_type, ?v1) \newline
triplet([canon], digicams.digital\_camera\_manufacturer.cameras, ?v0) \newline
answer(?v1) \newline

......(8 demonstrations omitted here.) \newline
% \#\#\#Question \newline
% what sensor\_resolution, jpeg ( exif 2.21 ) has cameras, cameras has sensor\_resolution, canon has cameras \newline
% \#\#\#PyQL \newline
% triplet([jpeg ( exif 2.21 )], digicams.camera\_compressed\_format.cameras, ?v0) \newline
% triplet(?v0, digicams.digital\_camera.sensor\_resolution, ?v1) \newline
% triplet([canon], digicams.digital\_camera\_manufacturer.cameras, ?v0) \newline
% answer(?v1) \newline

Please follow the format of exemplars and output PyQL query for the following question. No explanation or questioning allowed. \newline

Entity List: ['jpeg ( exif 2.21 )', 'canon'] \newline

\#\#\#Question \newline
the camera with a sensor from canon and a compression format of jpeg (exif 2.21) uses which viewfinder? \newline
\#\#\#PyQL  \\
\midrule

\textbf{LLM Output} & \textit{triplet([jpeg ( exif 2.21 )], digicams.camera\_compressed\_format.cameras, ?v0) \newline
triplet(?v0, digicams.digital\_camera.viewfinder\_type, ?v1) \newline
triplet([canon], digicams.camera\_sensor\_manufacturer.cameras, ?v0) \newline
answer(?v1)}  \\
\midrule
\textbf{Parsed SPARQL} & \textit{SELECT DISTINCT ?v1 WHERE \{\newline
:m.03h4lt3 :digicams.camera\_compressed\_format.cameras ?v0 . \newline
?v0 :digicams.digital\_camera.viewfinder\_type ?v1 . \newline
:m.01bvx1 :digicams.camera\_sensor\_manufacturer.cameras ?v0 .\newline
\}
}  \\
\midrule
\textbf{Execution Result} & \{m.01xrg1f: Optical, m.01xrg6z: TTL\} \\
\midrule
\textbf{F1 Score} & \textbf{1.0}  \\

\bottomrule
% \caption{A case study of \ours from GrailQA.}
\caption{A case study of \ours from GrailQA.}
\label{tab:case_study} \\
\end{longtable}
% }
\newpage

\begin{longtable}{@{}p{15.5cm}@{}}  % 调整列宽
% \caption{Prompt for KBQA.}
% \label{tab:show_prompt}\\
\toprule
\textbf{Prompt for KBQA}\\ 
\midrule
You are a powerful model for generating PyQL queries to answer natural language questions.
Here are some exemplars: \newline
\#\#\#Question \newline
what football\_league\_system, football\_league\_system has conference premier \newline
\#\#\#PyQL \newline
triplet(?v0, soccer.football\_league\_system.leagues, [conference premier])  \newline
answer(?v0) \newline

\#\#\#Question \newline
what leagues, football\_league\_system has conference premier, football\_league\_system has leagues \newline
\#\#\#PyQL \newline
triplet(?v0, soccer.football\_league\_system.leagues, [conference premier]) \newline
triplet(?v0, soccer.football\_league\_system.leagues, ?v1) \newline
answer(?v1) \newline

\#\#\#Question \newline
what sport, sport has conference premier \newline
\#\#\#PyQL \newline
triplet(?v0, sports.sport.leagues, [conference premier]) \newline
answer(?v0) \newline

% \#\#\#Question \newline
% what leagues, sport has conference premier, sport has leagues \newline
% \#\#\#PyQL \newline
% triplet(?v0, sports.sport.leagues, [conference premier]) \newline
% triplet(?v0, sports.sport.leagues, ?v1) \newline
% answer(?v1) \newline

% \#\#\#Question \newline
% what teams, sport has conference premier, sport has teams \newline
% \#\#\#PyQL \newline
% triplet(?v0, sports.sport.leagues, [conference premier]) \newline
% triplet(?v0, sports.sport.teams, ?v1) \newline
% answer(?v1) \newline

% \#\#\#Question \newline
% what sports\_team, league has conference premier, sports\_team has league \newline
% \#\#\#PyQL \newline
% triplet(?v0, sports.sports\_league\_participation.league, [conference premier]) \newline
% triplet(?v1, sports.sports\_team.league, ?v0) \newline
% answer(?v1) \newline
......(5 demonstrations omitted here.) \newline

% \#\#\#Question \newline
% what competitions\_of\_this\_type, competition\_type has conference premier, competition\_type has competitions\_of\_this\_type \newline
% \#\#\#PyQL \newline
% triplet(?v0, sports.sport.leagues, [conference premier]) \newline
% triplet(?v0, award.competition\_type.competitions\_of\_this\_type, ?v1) \newline
% answer(?v1) \newline

% \#\#\#Question \newline
% what sports\_official, sport has conference premier, sports\_official has sport \newline
% \#\#\#PyQL \newline
% triplet(?v0, sports.sport.leagues, [conference premier]) \newline
% triplet(?v1, sports.sports\_official.sport, ?v0) \newline
% answer(?v1) \newline

\#\#\#Question \newline
what positions, sport has conference premier, sport has positions \newline
\#\#\#PyQL \newline
triplet(?v0, sports.sport.leagues, [conference premier]) \newline
triplet(?v0, sports.sport.positions, ?v1) \newline
answer(?v1) \newline

\#\#\#Question \newline
what team\_coaches, sport has conference premier, sport has team\_coaches \newline
\#\#\#PyQL \newline
triplet(?v0, sports.sport.leagues, [conference premier]) \newline
triplet(?v0, sports.sport.team\_coaches, ?v1) \newline
answer(?v1) \newline

Please follow the format of exemplars and output PyQL query for the following question. No explanation or questioning allowed. \newline

Entity List: ['conference premier'] \newline

\#\#\#Question \newline
what are the names of the football leagues that are in the same football league system with conference premier? \newline
\#\#\#PyQL  \\

\bottomrule
\caption{Prompt for KBQA.}
\label{tab:show_prompt}\\
\end{longtable}

\begin{longtable}{@{}p{15.5cm}@{}}  % 调整列宽
% \caption{Prompts for Text2SQL.}
% \label{tab:show_prompt2} \\
\toprule
\textbf{Prompt for Text2SQL}\\ 
\midrule
You are a powerful model for generating SQL queries to answer natural language questions. \newline

Here are some related exemplars you can learn from:\newline

SELECT MIN(Magnitude (M bol )) FROM TABLE WHERE Radius (R \astrosun ) = '10'\newline

SELECT MIN(Mass (M \astrosun )) FROM TABLE WHERE Radius (R \astrosun ) = '10'\newline

SELECT MIN(Spectral type) FROM TABLE WHERE Radius (R \astrosun ) = '10'\newline

SELECT MIN(Star (Pismis24-\#)) FROM TABLE WHERE Radius (R \astrosun ) = '10'\newline

SELECT MIN(Radius (R \astrosun )) FROM TABLE WHERE Radius (R \astrosun ) = '10'\newline

SELECT MIN(Mass (M \astrosun )) FROM TABLE WHERE Star (Pismis24-\#) = '10'\newline

SELECT MIN(Temperature (K)) FROM TABLE WHERE Radius (R \astrosun ) = '10'\newline

SELECT MIN(Radius (R \astrosun )) FROM TABLE WHERE Star (Pismis24-\#) = '10'\newline

SELECT MIN(Magnitude (M bol )) FROM TABLE WHERE Star (Pismis24-\#) = '10'\newline

SELECT MIN(Spectral type) FROM TABLE WHERE Star (Pismis24-\#) = '10'\newline

Please output SQL query for the following question. No explanation or questioning allowed. Note that there is no need to use the LIKE keyword. And table name is TABLE for all questions.\newline

\#\#\#Question\newline
If a radius is 10, what is the lowest possible mass?\newline

\#\#\#Header\newline
['Star (Pismis24-\#)', 'Spectral type', 'Magnitude (M bol )', 'Temperature (K)', 'Radius (R \astrosun )', 'Mass (M \astrosun )']\newline

\#\#\#SQL \\
\bottomrule
\caption{Prompts for Text2SQL.}
\label{tab:show_prompt2} \\
\end{longtable}

\twocolumn

}

\end{document}